\documentclass{llncs}

\newif\ifdraft
\draftfalse

\usepackage[T1]{fontenc}
\usepackage[utf8]{inputenc}
\usepackage{lmodern}
\usepackage[scaled=.8]{beramono}
\usepackage{textcomp}
\usepackage{unicode-chars}
\usepackage{cmap}

\usepackage{xspace}

\ifdraft
\usepackage[show]{ed/ed}
\else
\usepackage[hide]{ed/ed}
\fi

\usepackage[hyperref=auto,firstinits=true]{biblatex}
\DeclareFieldFormat{labelnumberwidth}{#1.}
\usepackage[babel]{csquotes}
\MakeAutoQuote{“}{”}
\MakeAutoQuote*{‘}{’}
\usepackage{paralist}
\usepackage{graphicx}
\usepackage{lstsemantic}
\usepackage{url}

\newcommand{\Dolce}{\textmd{\textsc{Dolce}}\xspace}
\begin{document}

\title{The Distributed Ontology Language (DOL):\\ Use Cases, Syntax, and Extensibility\thanks{The development of DOL is supported by the German Research Foundation (DFG), Project I1-[OntoSpace] of the SFB/TR 8 “Spatial Cognition”; the first author is additionally supported by EPSRC grant EP/J007498/1.  The authors would like to thank Julian Kornberger and Henning Müller for implementing the Ontohub system, and the OntoIOp working group within ISO/TC 37/SC 3 for their feedback.}}
\author{%
Christoph Lange\inst{1,2}
\and Till Mossakowski\inst{1,3}
\and Oliver Kutz\inst{1} 
\and Christian Galinski\inst{4}
\and Michael Grüninger\inst{5}
\and Daniel Couto Vale\inst{1}
}
\institute{%
Research Center on Spatial Cognition, University of Bremen \email{christoph.lange@uni-bremen.de}, 
\email{okutz@informatik.uni-bremen.de}, \email{danielvale@gmail.com}
\and Computer Science, University of Birmingham, UK
\and DFKI GmbH Bremen \email{till.mossakowski@dfki.de}
\and Infoterm, Vienna \email{cgalinski@infoterm.org}
\and Department of Mechanical and Industrial Engineering, University of Toronto \email{gruninger@mie.utoronto.ca}
}

\maketitle

\begin{abstract}
  The Distributed Ontology Language (DOL) is currently being standardized within the OntoIOp (Ontology Integration and Interoperability) activity of ISO/TC 37/SC 3.  It aims at providing a unified framework for \begin{inparaenum}[(1)]
  \item ontologies formalized in heterogeneous logics,
  \item modular ontologies,
  \item links between ontologies, and
  \item annotation of ontologies.
  \end{inparaenum}
This paper presents the current state of DOL's standardization.  It focuses on use cases where distributed ontologies enable interoperability and reusability.  We demonstrate relevant features of the DOL syntax and semantics and explain how these integrate into existing knowledge engineering environments.
\end{abstract}

\section{Distributed Ontologies for Interoperability}
\label{sec:intr--distr}

An ontology is a formal description of the concepts and relationships that are of interest to an agent or a community of agents.  Today, ontologies are applied in eBusiness, eHealth, eGovernment, eInclusion, eLearning, smart environments, ambient assisted living (AAL), and virtually all other information-rich endeavours. An ontology facilitates semantic integration of knowledge and services in its application domain by providing a common model, onto which data from different sources, as well as descriptions of different services, can be mapped; thus, \emph{the ontology serves the goal of data and service interoperability}.

In complex applications, which involve multiple ontologies with overlapping concept spaces, data mapping is also required on a higher level of abstraction, viz.\ between different ontologies, and is then called ontology alignment.  While ontology alignment is most commonly studied for ontologies formalized in the same ontology language, the different ontologies used by complex applications may also be written in different ontology languages.  Popular choices include OWL, a language based on description logic (e.g.\ in biomedical applications and semantic web services) and Common Logic, a language based on first-order logic, which is required for formalizing mereology and notions of space and time, but exhibiting undecidable reasoning tasks.

Our approach faces this diversity not by proposing yet another ontology language that would subsume all the others.  Instead, we \emph{accept the diverse reality} of languages and tools optimized for them, and formulate means (on a sound and formal semantic basis) to \emph{compare and integrate ontologies that are written in different formalisms}.  We aim at addressing the challenge of checking the coherence (e.g.\ consistency %% OMITTED: IMHO too formal for TKE
%, conservativity,
or intended consequences) of ontologies and ontology-based services in a fully automated way.
  
Section~\ref{sec:distr-ontol-lang} gives a short overview of the OntoIOp standardization effort and the Distributed Ontology Language (DOL).  Section~\ref{sec:use-cases} introduces four use cases for this language.  Section~\ref{sec:syntax} explains its syntax, using examples from the use cases introduced previously.  Section~\ref{sec:registry} explains the mechanism for extending our framework by additional ontology languages, and section~\ref{sec:conclusion} concludes.

\section{The Distributed Ontology Language (DOL) – Overview}
\label{sec:distr-ontol-lang}

An ontology in the Distributed Ontology Language (DOL) consists of modules written in basic ontology languages, such as OWL or Common Logic.  These modules are serialized in the existing syntaxes of these languages as to facilitate reuse of existing ontologies.  DOL adds a meta-level on top, which allows for expressing heterogeneous ontologies and links between ontologies.\footnote{The languages that we call “basic” ontology languages here are usually limited to one logic and do not provide meta-theoretical constructs.}  Links can have a formal, logic-based semantics, as in the case of imports, conservative extensions (important for the study of ontology modules), and theory interpretations (important for reusing proofs), but there are also non-logical links, called alignments, with, e.g., statistical correspondences between ontology entities.  Thus, DOL gives ontology interoperability a formal grounding and makes heterogeneous ontologies and services based on them amenable to automated verification.

DOL is currently being standardized within the OntoIOp (Ontology Integration and Interoperability) activity of ISO/TC 37/SC 3\footnote{TC = technical committee, SC = subcommittee}.  The international working group comprises around 50 experts (around 15 active contributors so far), representing a large number of communities in ontological research and application, such as different \begin{itemize}
\item ontology languages and logics (e.g.\ Common Logic and OWL),
\item conceptual and theoretical foundations (e.g.\ model theory),
\item technical foundations (e.g.\ ontology engineering methodologies and linked open data), and
\item application areas (e.g.\ manufacturing).
\end{itemize}
For details and earlier publications, see the OntoIOp project page~\cite{OntoIOp}.

The OntoIOp/DOL standard is currently in the working draft stage and will be submitted as a committee draft (the first formal ISO standardization stage) in August 2012.\footnote{The standard draft itself is not %currently
publicly available, but negotiations are under way to make the final standard document public, as has been done with the related Common Logic standard~\cite{CommonLogic:biblatex}.}  The final international standard ISO 17347 is scheduled for 2015.

The standard specifies syntax, semantics, and conformance criteria:
\begin{description}
\item[Syntax:] the abstract syntax of distributed ontologies and their parts (basic ontologies and links), as well as three concrete syntaxes (serializations) detailed in section~\ref{sec:syntax}: a text-oriented one for humans, XML and RDF for exchange among tools and services, where RDF particularly addresses exchange on the Web.
\item[Semantics:] The three alternative formal semantics, which are compatible with each other, address different reusability and verification requirements:
  \begin{description}
  \item[Direct set-theoretical semantics:] covers the core of the language, and is extended by an \emph{institutional and category-theoretic semantics} for advanced features such as ontology combinations%% OMITTED for TKE
% (technically co-limits)
.  Basic ontologies keep the original semantics of their languages.
  \item[Translational semantics:] employs the semantics of the expressive Common Logic ontology language for all basic ontologies, taking advantage of the fact that for all basic ontology languages known so far translations to Common Logic have been specified or are known to exist
  \item[Collapsed semantics:] This is a third option, which has not yet been realized.  The semantics of the meta-theoretical language level provided by DOL (logically heterogeneous ontologies and links between them) is not just specified on paper in semiformal mathematical textbook style, but once more formalized in Common Logic, thus in principle allowing for machine verification of meta properties.\end{description}
For details about the formal semantics, see~\cite{MLK:3SemanticsForDistributedOntologies12}.  This paper addresses semantics from the perspective of integrating new logics and logic translations into the OntoIOp framework (cf.\ section~\ref{sec:registry}).
\item[Conformance criteria] provide for DOL's extensibility to other basic ontology languages than those considered so far, including possible future languages.
  \begin{description}
  \item[A basic ontology language] conforms with DOL if its underlying logic has a set-theoretic or, for the extended DOL features, an institutional semantics.  Similar criteria apply to translations between languages.
  \item[A serialization] of a basic ontology language conforms if it supports IRIs for Web-scalable identification of entities and satisfies some further well-formedness criteria.
  \item[A document] conforms if it is well-formed w.r.t.\ one of the DOL serializations and one conforming serialization of each basic ontology language used.  This particularly requires explicitly mentioning all logics and translations employed.
  \item[An application] essentially conforms if it is capable of processing conforming documents, and providing logical information that is implied by the formal semantics.
  \end{description}
\end{description}

\section{Use Cases}
\label{sec:use-cases}

As DOL is still in an early stage of development, we can not yet demonstrate real-world application settings where the use of DOL made a difference w.r.t.\ knowledge integration and service interoperability.  However, we are working on basic tools that support processing the DOL syntax and semantically verifying distributed ontologies (cf.\ section~\ref{sec:ontohub}), and we have started to apply them to the verification of meta-theoretical relationships in a collection of generic ontologies (section~\ref{sec:colore}), to representing alignments and multilinguality (section~\ref{sec:do-roam}).  With DOL being based on existing ontology standards and designed with existing ontology engineering practices in mind, we are aware of a number of existing interoperability use cases in which DOL will be able to improve application support.  This up-front introduction of use cases prepares for a detailed discussion of how DOL's syntax and semantics afford application support in the respective settings.

\subsection{Ontohub, a repository engine for managing distributed ontologies}
\label{sec:ontohub}

The Open Ontology Repository (OOR) initiative aims at ``promot[ing] the global use and sharing of ontologies by \begin{inparaenum}[(i)]\item establishing a hosted registry-repository; \item enabling and facilitating open, federated, collaborative ontology repositories, and \item establishing best practices for expressing interoperable ontology and taxonomy work in registry-repositories\end{inparaenum}, where an ontology repository is a facility where ontologies and related information artifacts can be stored, retrieved and managed''~\cite{OOR:webpage}.  OOR is a long-term initiative, which has not resulted in a complete implementation so far, but established requirements and designed an architecture\footnote{See \url{http://ontolog.cim3.net/cgi-bin/wiki.pl?OpenOntologyRepository_Requirement} and \url{http://ontolog.cim3.net/cgi-bin/wiki.pl?OpenOntologyRepository_Architecture}, respectively}.  Our Ontohub repository engine~\cite{Ontohub} aims at satisfying a subset of the OOR requirements, with a particular focus on managing distributed heterogeneous ontologies.  

\begin{figure}
  % actual width: 625
  \begin{minipage}[c]{.29\textwidth}\vspace{0pt}\includegraphics[width=\textwidth]{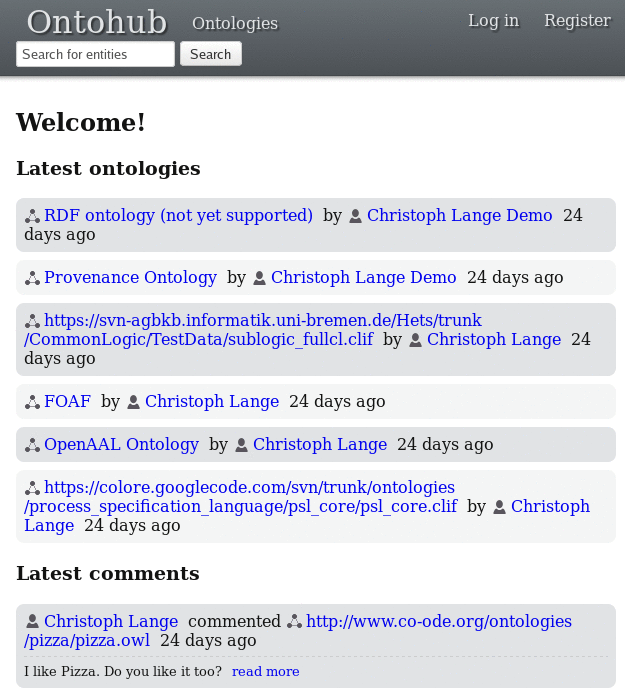}\end{minipage}%
  \hspace{.005\textwidth}%
  % actual width: 622
  \begin{minipage}[c]{.29\textwidth}\vspace{0pt}\includegraphics[width=\textwidth]{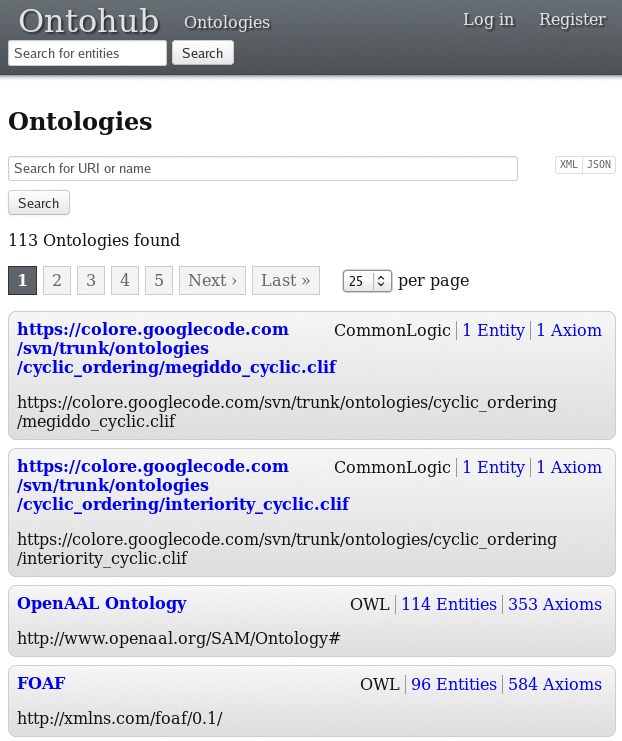}\end{minipage}%
  \hspace{.005\textwidth}%
  % actual width: 877
  \begin{minipage}[c]{.41\textwidth}\vspace{0pt}\includegraphics[width=\textwidth]{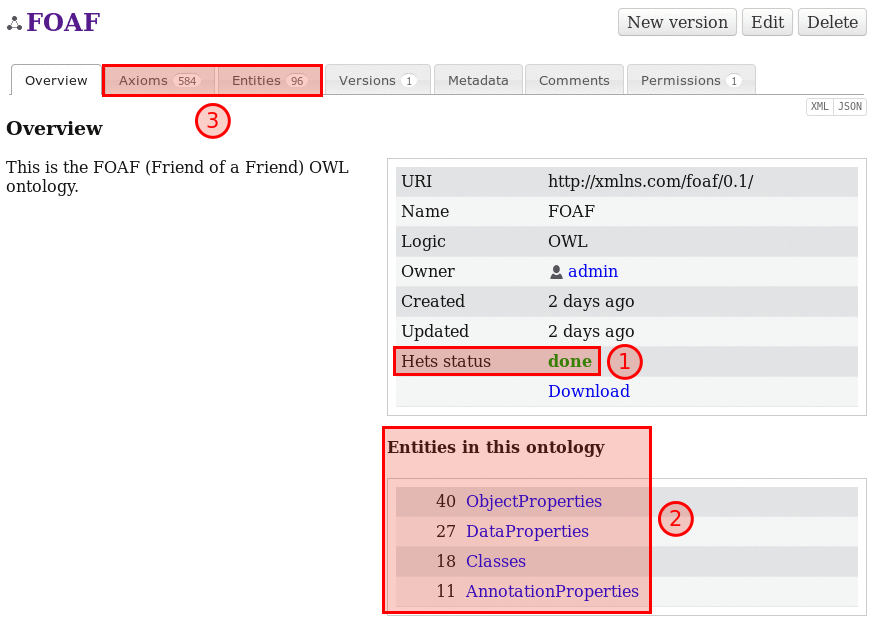}\end{minipage}%
  \caption{The Ontohub web frontend: welcome page, ontology overview, view of one basic ontology}
  \label{fig:ontohub}
\end{figure}
\begin{figure}
  \centering
  \includegraphics[width=\textwidth]{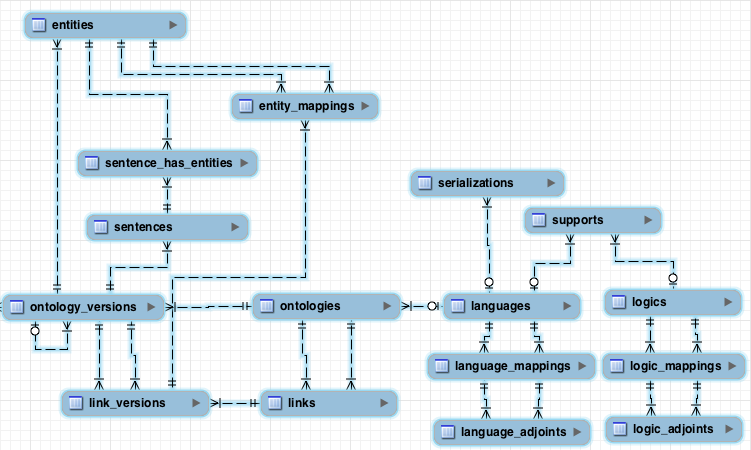} 
  \caption{Subset of the Ontohub database schema (entity-relationship diagram using crow's foot notation); 
  left side: ontologies; right side: OntoIOp registry (cf.\ section~\ref{sec:registry})}
  \label{fig:ontohub-db}
\end{figure}

Users of Ontohub can upload, browse, search and annotate basic ontologies in various languages via a web frontend (cf.\ figure~\ref{fig:ontohub}).  Ontohub accesses the Heterogeneous Tool Set (Hets~\cite{HETS:on,HETS:pub}) via a RESTful web service interface for having the structure of ontologies analyzed.  Hets supports a large number of basic ontology languages and logics, and is capable of describing the structural outline of an ontology from the perspective of DOL, which is not committed to one particular logic.  From this perspective, a basic ontology consists of \emph{sentences} (e.g.\ axioms), \emph{entities} (e.g.\ classes, properties or individuals in OWL, or names and sequence markers in Common Logic), and, if supported by the respective language, \emph{imports} of other ontologies.  This structural information is stored in the Ontohub database and exposed to human users via a web interface and to machine clients as RDF linked data~\cite{HB:LinkedData11}.  Beyond basic ontologies, Ontohub supports linking ontologies, across ontology languages, and creating distributed ontologies as sets of basic ontologies and links among them, as can be seen from the left half of the diagram in figure~\ref{fig:ontohub-db}, which closely corresponds to the abstract syntax of DOL.
Note that the Ontohub database schema takes advantage of another useful abstraction: Same as basic ontologies, we treat distributed ontologies as ontologies.  The entities of distributed ontologies are ontologies (basic, or, in complex scenarios, again distributed), and their sentences are links.

\subsection{Verifying meta-theoretical relationships in COLORE}
\label{sec:colore}

COLORE, the Common Logic Repository, is an open repository of more than 500 Common Logic ontologies.  The objective of COLORE is to provide an “adequate set of generic ontologies that can be used to specify the semantics of primitive concepts”, as, for example, “any product ontology must refer to relationships from geometry and topology, and different manufacturing standards may require different ontologies for time”\footnote{\url{http://colore.googlecode.com}}.  One of the primary applications of COLORE is to support the verification of ontologies for commonsense domains such as time, space, shape, and processes.  Verification consists in proving that the ontology is equivalent to a set of core ontologies for mathematical domains such as orderings, incidence structures, graphs, and algebraic structures.   COLORE comprises core ontologies that formalize algebraic stuctures (such as groups, fields, and vector spaces), orderings (such as partial orderings, lattices, betweenness), graphs, and incidence structures in Common Logic, and, based on these, representation theorems for generic ontologies for the above-mentioned commonsense domains.

COLORE stores metadata about its ontologies, which are represented using a custom XML schema that covers the following aspects\footnote{\url{http://stl.mie.utoronto.ca/colore/metadata.html}}, without specifying a formal semantics for them:
\begin{description}
\item[Module provenance:] author, date, version, description, keyword, and parent ontology
\item[Axiom source provenance:] name, author, and date
\item[Direct relations:] maps (i.e.\ signature morphisms), definitional extension, conservative extension (important for the study of ontology modules), inconsistency between modules, imports, relative interpretation and faithful interpretation (important for reusing proofs), and definable equivalence
\end{description}

DOL provides built-in support for a subset of the ``direct relations'' and specifies a formal semantics for them.  We have started to automatically verify COLORE's meta-theoretical relationships using Hets~\cite{HETS:on,HETS:pub}.  In addition, DOL allows for implementing the remainder of the COLORE metadata vocabulary as an ontology, reusing suitable existing metadata vocabularies such as OMV~\cite{OMV:web,HPSSHGS05:OMV}, and it allows for implementing one or multiple Common Logic ontologies plus their annotations as one coherent distributed ontology.

\subsection{Aligning tag ontologies and generating multilingual labels in the DO-ROAM interface}
\label{sec:do-roam}

The DO-ROAM (\textbf{D}ata and \textbf{O}ntology driven \textbf{R}oute-finding \textbf{O}f \textbf{A}ctivity-oriented \textbf{M}obility) route-finding web service\footnote{\url{http://www.do-roam.org}} is driven by a set of aligned OWL ontologies by which places on a map are tagged~\cite{CodescuHKMR11}.  Its map-based user interface offers multilingual labels, which are maintained in close connection to the ontologies.  The ongoing port of DO-ROAM to DOL allows for coherently representing the aligned ontologies as one distributed ontology, including alignments determined by, e.g., the Falcon matching tool~\cite{HuQu-08}.  It will also ease the maintenance of the user interface labels by allowing to keep them as annotations inside the distributed ontology.%% OMITTED: This doesn't really apply here.
%, taking advantage of DOL's ability to also identify and annotate those parts of an OWL ontology whose identification or annotation OWL itself does not support.

\subsection{Connecting devices of differing complexity in an Ambient Assisted Living setting}
\label{sec:aal}

The OntoIOp activity was initially motivated by interoperability needs in ambient assisted living (AAL)~\cite{KMGL:HetOntoInteropStandard11}.  AAL environments involve a number of different devices that need to interact with their users, often elderly persons or persons with disabilities, with each other, and possibly with a central home controller.  Existing AAL ontologies, such as the OpenAAL\footnote{\url{http://openaal.org}} OWL ontology, usually do not cover AAL scenarios in the full width and depth required. For example:
\begin{description}
\item[Width (of the domain):] OpenAAL covers assisted persons and assistive devices, including kitchen devices, but not nutrition requirements.  The latter knowledge needs to be imported from other ontologies, as well as from linked open datasets on the Web, as far as ingredients of concrete products are concerned.
\item[Depth (of the representation):] Different devices have different complexities.  The behavior of a light switch can be described in propositional logic; describing it in OWL may exceed the computational resources available.  On the other hand, a spatial calculus, which a smart wheelchair needs for navigation, requires full first order logic.  Thus, the integration of different devices requires a logically \emph{heterogeneous} ontology.
\end{description}
DOL is capable of capturing all information that is relevant in a complex AAL scenario within one distributed ontology of heterogeneous modules arranged around the OWL core of OpenAAL, including logical links between OpenAAL and the other ontologies.

\section{Syntax}
\label{sec:syntax}

This section shows concrete examples of the DOL syntax.  The examples use the DOL Text serialization, which is designed for human authors, but we also briefly describe the specific characteristics of the XML and RDF serializations.  We start with a heterogeneous formalization of mereology as a general example that covers large parts of DOL, but then also point out specific aspects in the context of the use cases introduced previously.

\subsection{Generic example: heterogeneous formalization of mereology}
\label{sec:mereology}

Mereology lends itself well to a logically heterogeneous formalization.  Mereological relations such as parthood are frequently \emph{used} in ontologies (e.g.\ biomedical ontologies), but many of these ontologies are formalized in languages that are not fully capable of \emph{defining} the mereological notions (e.g.\ the EL profile of OWL, designed for efficient reasoning with a large number of entities – a frequent case in the biomedical domain).  Listing~\ref{lst:example} shows that propositional logic is already capable of describing the basic categories, over which the \Dolce foundational ontology~\cite{WonderWeb:OntologyLibrary} defines mereological relations.  Propositional logic is popular in formal modelling, since consistency and logical consequence can be decided efficiently, which allows for early detection of modeling errors in an overall ontology design.  The same knowledge can be formalized more conveniently in OWL, which additionally allows for describing basic parthood properties.  As our OWL ontology redeclares the same categories as the propositional logic ontology \textit{Taxonomy}, just using different names, we observe that it \emph{interprets} the former, which tools such as the above-mentioned Hets can prove by translating \textit{Taxonomy} to OWL and verifying that all of its axioms, after translation, also hold in the OWL ontology.  Finally, we provide a full definition of several mereological relations in first order logic, in Common Logic, importing, translating and extending the OWL ontology.  Common Logic extends first-order logic with some second-order features.  We use the possibility to quantify over predicates to concisely express the restriction of the variables $x$, $y$, and $z$ to the same taxonomic category.

\begin{lstlisting}[label={lst:example},caption={A heterogeneous ontology for mereology~\cite{MLK:3SemanticsForDistributedOntologies12,KML:Hyperontologies10}},basicstyle=\ttfamily\scriptsize,language=dolText,morekeywords={props,ObjectProperty,Class,DisjointUnionOf,SubClassOf,Characteristics,Transitive,Asymmetric,SubPropertyOf,DisjointClasses,EquivalentTo,inverse,only,forall,iff,if,or,exists},escapechar=@,mathescape]
%prefix( :      <http://www.example.org/mereology#>             %% prefix for this distributed ontology
         owl:   <http://www.w3.org/2002/07/owl#>                         %% OWL basic ontology language
         log:   <http://purl.net/dol/logics/>                         %% DOL-conforming logics@\itshape\ (Fig.~\ref{fig:graph})@
         trans: <http://purl.net/dol/translations/>                %% translations between these logics
         ser:   <http://purl.net/dol/serializations/> %)      %% serializations, i.e. concrete syntaxes

distributed-ontology Mereology

logic log:Propositional syntax ser:Prop/Hets              %% non-standard serialization built into Hets
ontology Taxonomy =                    %% basic taxonomic information about mereology reused from DOLCE
  props PT, T, S, AR, PD
  . S $\vee$ T $\vee$ AR $\vee$ PD $\longrightarrow$ PT                                                           %% PT is the top concept
  . S $\wedge$  T  $\longrightarrow$ $\bot$                                                      %% PD, S, T, AR are pairwise disjoint
  . T $\wedge$ AR $\longrightarrow$ $\bot$                                                                                %% and so on

logic log:SROIQ syntax ser:OWL2/Manchester                                     %% OWL Manchester syntax
ontology BasicParthood =                                    %% Parthood in SROIQ, as far as expressible
  Class: ParticularCategory SubClassOf: Particular %% omitted similar declarations of the other classes
    DisjointUnionOf: SpaceRegion, TimeInterval, AbstractRegion, Perdurant
                                        %% pairwise disjointness more compact thanks to an OWL built-in
  ObjectProperty: isPartOf        Characteristics: Transitive
  ObjectProperty: isProperPartOf  Characteristics: Asymmetric  SubPropertyOf: isPartOf
  Class: Atom EquivalentTo: inverse isProperPartOf only owl:Nothing

interpretation TaxonomyToParthood : Taxonomy to BasicParthood =
  translate with trans:PropositionalToSROIQ,           %% translate the logic, then rename the entities
  PT $\mapsto$ Particular, S $\mapsto$ SpaceRegion, T $\mapsto$ TimeInterval, A $\mapsto$ AbstractRegion, %[ and so on ]%

logic log:CommonLogic syntax ser:CommonLogic/CLIF %% syntax: the Lisp-like CLIF dialect of Common Logic
ontology ClassicalExtensionalParthood =
  BasicParthood translate with trans:SROIQtoCL  %% import the OWL ontology from above, translate it ...
  then {                                                  %% ... to Common Logic, then extend it there:
    (forall (X) (if (or (= X S) (= X T) (= X AR) (= X PD))
                    (forall (x y z) (if (and (X x) (X y) (X z))
                                        (and                                 %% now list all the axioms
      (if (and (isPartOf x y) (isPartOf y x)) (= x y))                                  %% antisymmetry
      (if (and (isProperPartOf x y) (isProperPartOf y z)) (isProperPartOf x z))
                                     %% transitivity; can't be expressed in OWL together with asymmetry
      (iff (overlaps x y) (exists (pt) (and (isPartOf pt x) (isPartOf pt y))))
      (iff (isAtomicPartOf x y) (and (isPartOf x y) (Atom x)))
      (iff (sum z x y) (forall (w) (iff (overlaps w z) (and (overlaps w x) (overlaps w y)))))
      (exists (s) (sum s x y)))))))                                             %% existence of the sum
  }
\end{lstlisting}

\subsection{Web-scalable identification with IRIs}
\label{sec:iris}

All DOL serializations use IRIs for web-scalable, Unicode-aware identification of ontologies and all of their parts.  DOL Text and DOL XML employ CURIEs (compact URI expressions~\cite{w3c:rdfa-core}) for abbreviating long IRIs, whereas DOL RDF uses the similar abbreviation facilities provided by the actual RDF serialization used (e.g.\ Turtle or RDF/XML).  For example, the initial declaration of the “empty prefix” causes the distributed ontology's identifier to expand to the full IRI \url{http://www.example.org/mereology#Mereology}.  The prefixes defined on the level of the distributed ontology propagate into the basic ontologies, unless they are overridden there; in particular, the empty prefix applies to all unprefixed identifiers having a global scope.  This is particularly useful for dealing with basic ontology languages that do not support IRIs as identifiers, as the propositional language used here, or that do not enforce their use, as Common Logic.  In our example, this mechanism places the entities of all basic ontologies in the same IRI namespace; external resources would be able to refer to these entities using IRIs such as \url{http://www.example.org/mereology#isAtomicPartOf}.

The consistent use of IRIs in distributed ontologies enables publishing them as \emph{linked data}~\cite{HB:LinkedData11}.  This means: When the DOL file in listing~\ref{lst:example}, or, preferably, a machine-friendy variant in the DOL RDF serialization, is made available under the URL \url{http://www.example.org/mereology}, machine clients can retrieve a description of our distributed ontology and any of its parts (and, by choice of the IRIs of the entities in the basic ontology, also of these entities) by simply dereferencing their IRIs, i.e.\ treating them as URLs.

\subsection{Expressing complex interpretations (COLORE use case)}
\label{sec:interpretations}

Expressing the meta-theoretical relationships of COLORE (introduced in section~\ref{sec:colore}) in DOL requires more effort than shown in the previous example.  Listing~\ref{lst:example} shows an example\footnote{An excerpt from \url{https://colore.googlecode.com/svn/trunk/ontologies/complex/owltime/owltime_interval/mappings/owltime_le.dol}; the individual ontologies are actually stored in separate files, but here we demonstrate DOL's ability to maintain different ontologies within one file.} for interpreting linear orders (\url{linear_ordering}) as orders between time intervals that begin and end with an instant (\url{owltime_le}).  A third ontology (\url{mappings/owltime2orderings}) takes care of mapping the different predicate names used by the source and the target ontology, respectively.  This mapping ontology is implemented separately to facilitate maintenance and to enable its reuse in different related mapping scenarios.
 
We state that the source ontology can be interpreted in terms of the union of the target ontology and the mapping ontology in a model-theoretically conservative way, and that \url{mappings/owltime2orderings} extends
\url{owltime_le} with definitions.  

\begin{lstlisting}[caption={Interpreting linear orders as orders between time intervals},label={lst:colore},basicstyle=\ttfamily\scriptsize,language=dolText,morekeywords={props,ObjectProperty,Class,DisjointUnionOf,SubClassOf,Characteristics,Transitive,Asymmetric,SubPropertyOf,DisjointClasses,EquivalentTo,inverse,only,cl-imports,forall,iff,if,or,exists},escapechar=@,mathescape]
%prefix(                                    %% log: and ser: prefix declarations omitted; see @\itshape listing~\ref{lst:example}@
 :    <http://code.google.com/p/colore/.../owltime/owltime_interval/mappings/owltime_le.dol#>
 int: <http://code.google.com/p/colore/.../owltime/owltime_interval/>   %% namespaces of the ontologies
 ord: <http://code.google.com/p/colore/.../orderings/> )%               %% in this distributed ontology

logic log:CommonLogic syntax ser:CommonLogic/CLIF

ontology ord:linear_ordering =    %% here using Common Logic's import facility as an alternative to ...
  (cl-imports ord:partial_ordering) (forall (x y) (or (leq x y) (leq y x) (= x y)))

%% ... to DOL's general one: We create the ontology of linearly ordered time intervals that ...
ontology int:owltime_le = int:owltime_linear then int:owltime_e  %% ... begin and end with an instant 
%% ... by extending linearly ordered time intervals with intervals that begin and end with an instant

ontology int:mappings/owltime2orderings = (forall (x y) (iff (leq x y) (or (before x y) (= x y))))
  (forall (x y) (iff (lt x y) (before x y)))          %% map time intervals to general linear orderings

interpretation i %mcons :                      %% interpreting linear orderings as time interval orders
  ord:linear_ordering to {int:owltime_le and %def int:mappings/owltime2orderings}
\end{lstlisting}

\subsection{Complex alignments, and multilingual translation of concept labels (DO-ROAM use case)}
\label{sec:alignment-multilingual}

In the DO-ROAM use case introduced in section~\ref{sec:do-roam}, we align an ontology of activities with an ontology of OpenStreetMap tags.  Concepts from the activities ontology (listing~\ref{lst:do-roam} shows places) do not always correspond to simple tags but can also correspond to a complex combination of tags, which we express by mapping entities to OWL terms.

For use on the user interface, the concept labels have been translated to multiple natural languages; for example, the German label of the concept \textit{ChargingStation} is “Ladestation”.  The data structures describing these translations are given as YAML\footnote{\url{http://yaml.org}} datasets, one per language.   As the expressivity of these datasets is similar to that of RDF, we treat, for the purpose of DO-ROAM, YAML as a serialization of the RDF ontology language.

\begin{lstlisting}[caption={Complex expressions in alignments, and including external resources for translating concept labels},label={lst:do-roam},basicstyle=\ttfamily\scriptsize,language=dolText,morekeywords={a,Individual,exactly,props,ObjectProperty,Class,DisjointUnionOf,SubClassOf,Characteristics,Transitive,Asymmetric,SubPropertyOf,DisjointClasses,EquivalentTo,inverse,only,cl-imports,forall,iff,if,or,exists},escapechar=@,mathescape]
%prefix(
  do-roam: <https://raw.github.com/doroam/planning-do-roam/master/>
  activ:   <https://raw.github.com/doroam/planning-do-roam/master/Ontology/activities.owl#>
  tags:    <https://raw.github.com/doroam/planning-do-roam/master/Ontology/tags.owl#>
)%

language lang:OWL2/DL

alignment do-roam:ActivitiesToTags : activ: to tags: =
  activ:Restaurant = $\exists$ tags:has_k_amenity . tags:v_restaurant,
  %% "=" is equivalence as defined in the Alignment API @\cite{AlignmentAPI,DEST:AlignmentAPI11}@
  activ:ChargingStation =          
    $\exists$ tags:has_k_amenity . tags:v_charging_station
           $\sqcup$ ($\exists$ tags:has_k_amenity . tags:v_fuel $\sqcap\ \exists$ tags:has_k_fuel:electricity . tags:yes),
    %% We give OWL complex OWL class expressions in German DL notation for brevity; actually one would 
    %% use, e.g., OWL Manchester Syntax here.
  ...

ontology do-roam:ActivityTranslation = 
  activ: project with proj:OWL2DLtoRDF                         %% project activity ontology down to RDF
    then language lang:RDF syntax ser:RDF/YAML :          %% add German translations as RDF from a file
      do-roam:config/locales/de.yml                                %% serialized in (non-standard) YAML
\end{lstlisting}
% then language lang:RDF syntax ser:RDF/Turtle :
%   do-roam:config/locales/translations.rdf           %% add translations in RDF serialized as Turtle
\ednote{TM@CL: I am not sure how to integrate the YAML files. Should we say
that YAML is a serialization of RDF? Maybe we should integrate true RDF files instead (see ``alternatively'')--- they do not exist yet, but would contain triples like (translation-de cinema Kino) and (translation-instance de translation-de). Here, cinema is a concept in do-roam:Ontology/activity.owl. In a sense, the triple containing the translation for cinema is not directly related to the concept cinema. However, probably a tool would be satisfied with this and fetch the needed triples when displaying the translation to a user.}

\subsection{Reusing linked open datasets (AAL use case)}
\label{sec:lod}

The AAL use case introduced in section~\ref{sec:aal} requires ontology-based services to know not only general nutrition requirements but also the ingredients of concrete products.  For a concrete example, we reuse the Pizza OWL ontology~\cite{PizzaOntology} for general descriptions of types of pizzas, and the ProductDB linked open dataset\footnote{\url{http://productdb.org}} for descriptions of products (manufacturer, name, and identifiers).  Our own distributed ontology connects the former ontologies by redeclaring a product of ProductDB as an instance of a specific type of pizza and asserting some properties about its ingredients.  We express this set of assertions (in description logic terminology: the ABox) in two parts, an RDF and an OWL DL part, which thus become immediately accessible to reasoning tools specialized for these two ontology languages.

\begin{lstlisting}[caption={Connecting ontologies and linked open datasets about nutrition},label={lst:lod},basicstyle=\ttfamily\scriptsize,language=dolText,morekeywords={a,Individual,exactly,props,ObjectProperty,Class,DisjointUnionOf,SubClassOf,Characteristics,Transitive,Asymmetric,SubPropertyOf,DisjointClasses,EquivalentTo,Types,inverse,only,cl-imports,forall,iff,if,or,exists},escapechar=@,mathescape]
%prefix( pizza: <http://www.co-ode.org/ontologies/pizza/pizza.owl#>
         productdb: <http://productdb.org/ean/> )%

language lang:OWL2/DL                      %% here: characterizing ontologies by language, not by logic
ontology FreezerInMyHome = 
  { pizza: project with proj:OWL2DLtoRDF                 %% Project Pizza OWL 2 DL ontology down to RDF
    and productdb:                                                 %% Merge with ProductDB RDF ontology
    then language lang:RDF syntax ser:RDF/Turtle : {                    %% Add the RDF part of our ABox
         %% Talk about a concrete pizza.  This item doesn't actually exist in the ProductDB linked open
         %% dataset, but we pretend so.
      productdb:4001724819806 pizza:hasTopping
        [ a pizza:TomatoTopping ], [ a pizza:MozzarellaTopping ] .
  } translate with trans:RDFtoOWL2DL
  then { pizza:                                    %% Pizza ontology once more, now keep it in OWL 2 DL
    then syntax ser:OWL2/Manchester : {                                 %% Add the OWL part of our ABox
    Individual: productdb:4001724819806            %% Redeclare the concrete pizza as an OWL individual
      %% We need OWL for expressing that besides tomato and mozzarella this pizza does not have any
      %% other toppings (and note that the two toppings of our concrete pizza are assumed to be
      %% different because the Pizza ontology declares TomatoTopping and MozzarellaTopping disjoint).
      Types: pizza:hasTopping exactly 2
  }
\end{lstlisting}

Note that linked datasets are just a special case of ontologies, usually in the logic RDF, which may require special treatment due to their large size.  Most commonly, linked datasets are published as one record of RDF per resource in the dataset, each one having links to related resources.  Our current implementation of the Ontohub and Hets systems introduced in section~\ref{sec:ontohub} does not yet support this way of deployment but insteads expects ontologies to be downloadable as single files.

\section{Extensibility: A Registry for Ontology Languages and Mappings}
\label{sec:registry}

The examples in the previous sections demonstrated different ontology languages, logics and serializations supported by them, and mappings (translations or projections) between logics and ontology languages.  The OntoIOp standard is not limited to a fixed set of ontology languages.  It will rather be possible to use any (future) ontology language, logic, serialization, or mapping with DOL, once its conformance with the criteria specified in the standard has been established.  Any such resource shall be identified by an IRI, so that DOL ontologies can refer to it.  At these IRIs there shall be a machine-readable description of the resource according to the linked data principles (at least in RDF, but see below), so that, for example, any agent given a basic ontology can find out the ontology languages this ontology can be translated into.  We have realized the RDF vocabulary for these descriptions as a subset of the ontology that implements the DOL RDF serialization.\footnote{The namespace IRI of this ontology is \url{http://purl.net/dol/1.0/rdf#}.}

\begin{figure}
  \centering
  \includegraphics[width=\textwidth]{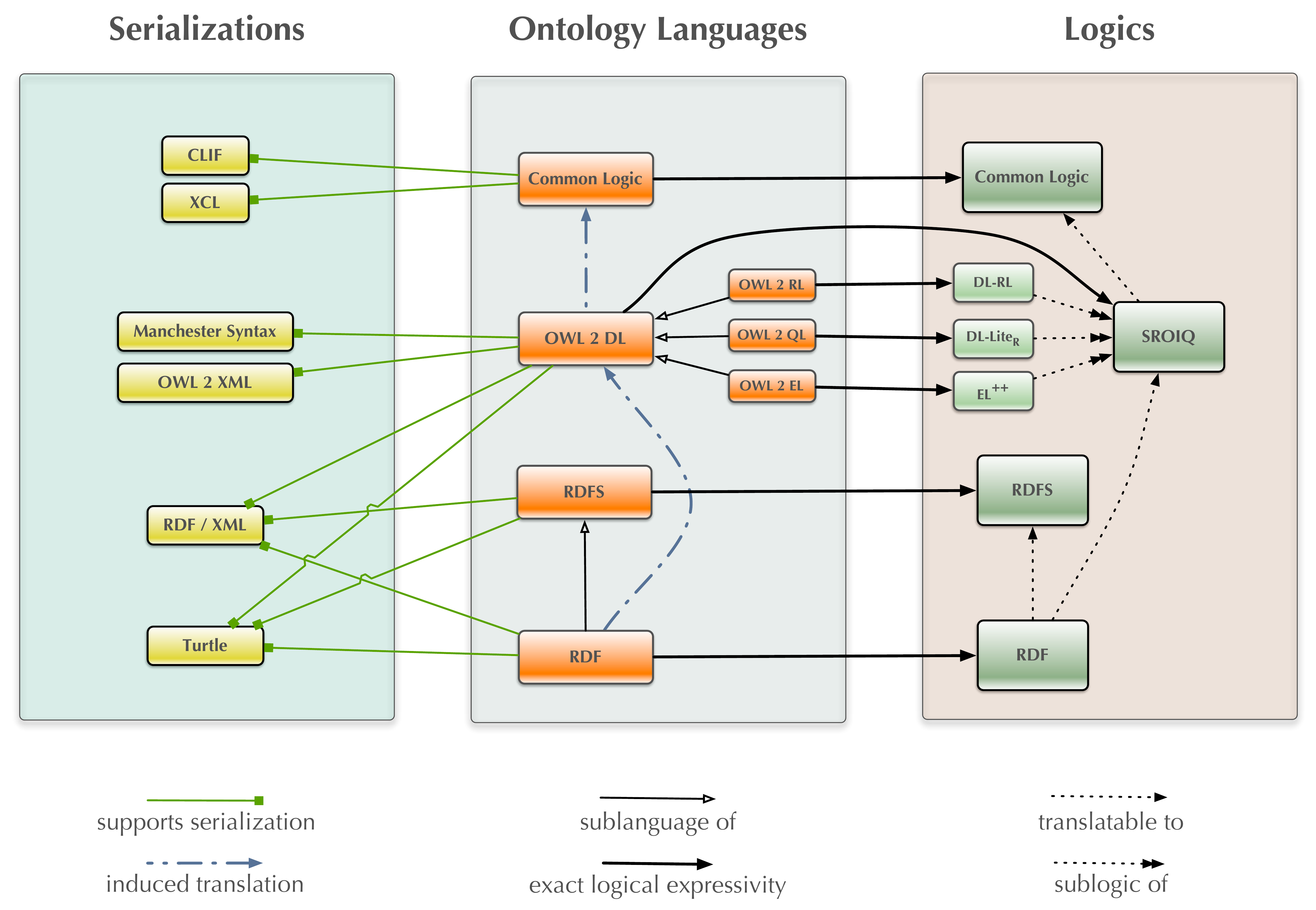} 
  \caption{Subset of the OntoIOp registry, shown as an RDF graph}
  \label{fig:graph}
\end{figure}

The IRIs of the resources mentioned so far will be recorded in a central registry.  In the current, early phase of the OntoIOp standardization process, \emph{we} are maintaining this registry.  With the release of the final international standard, everyone will be able to make contributions, which an editorial board will review and approve or reject.  The registry is, and will be, hosted in a dedicated installation of the Ontohub repository engine introduced in section~\ref{sec:ontohub}; compare the right half of the diagram in figure~\ref{fig:ontohub-db} to figure~\ref{fig:graph}.

Figure~\ref{fig:graph} shows a subset of the current registry.\footnote{As an entry point for exploring the linked dataset, take, e.g., the OWL 2 DL language, whose IRI is \url{http://purl.net/dol/languages/OWL2/DL}; for user-friendly browsing frontends, see \cite{HB:LinkedData11,DR:VisualisingLinkedData11}}  The central concept is the ontology language, whose semantics is usually defined by a logic.  Storing and exchanging an ontology requires writing it down in one of the serializations supported by its ontology language.  The separation into ontology languages and logics is necessary for a reason not obvious from figure~\ref{fig:graph}: There are ontology languages whose exact logical expressivity is not known (e.g.\ the biomedical ontology language OBO before version 1.4~\cite{OBO1.4}), and there are logics that do not directly correspond to an ontology language but that are the source or the target of translations to or from logics that do (e.g.\ many-sorted first-order logic~\cite{MoKu:OntoGraph11}), and therefore are relevant for OntoIOp.  DOL provides constructs for declaring the language, logic, and serializations of the subsequent ontologies; in most cases, it suffices to declare two of them, from which the third can be inferred from the registry.

Most mappings (translations or projections) between ontology languages are backed by corresponding mapping between logics.  Figure~\ref{fig:graph} shows some translations.  For most logic translations (from a less expressive logic to a more expressive one), one can specify an adjoint \emph{projection} (from the more expressive logic to the less expressive one, not shown here).  Note that there can be multiple alternative translations between a pair of logics; if more than one is in the registry, DOL selects a default translation, unless the desired translation is specified explicity (see \cite{MLK:3SemanticsForDistributedOntologies12} for a detailed discussion).  If an ontology language translation is backed by a logic translation, DOL allows for referring to it either way, using the IRI of the ontology language translation or of the logic translation; again, a lookup in the registry will help to disambiguate.

Finally we remark that we require ontology languages, logics, translations, etc., to be \emph{described} at least in RDF.  In this case, the full definition would be given in a specification document in textbook style.  But we actually encourage machine-comprehensible exhaustive \emph{definitions} of these resources, which can be published in parallel to the RDF descriptions at the same IRIs = URLs, where clients would request the desired content by HTTP content negotiation.  Such definitions of most logics and logic translations have been implemented in the OMDoc/MMT language in the Logic Atlas project~\cite{CodHorKoh:palai11}.

\section{Conclusion}
\label{sec:conclusion}

% \ednote{CL@TM: A possible future direction: our idea of “Extending OWL with datatypes defined in CASL” is also (or rather \emph{alternatively}) worth elaborating here?  My intuition is that terminology deals a lot with what RDF calls “literals”, and datatypes help to get more semantics into them.  OTOH OWL's support for custom XML Schema datatypes (which not all OWL \emph{reasoners} support though) is possibly sufficient for terminology applications (?).}
Integration and interoperability of ontological knowledge, i.e.\ reusability, embeddability into services and devices, and repurposability across application settings must be based on
\begin{enumerate}%[(1)]
\item consistent methodology standards for data models and data modelling,
\item coordinated standardisation of several kinds of structured content,
\item standardized identification systems for individual pieces of information,
\item standardized transfer protocols and interchange formats, in order to be efficient and reliable,
\item standardized metamodels and a standardized meta ontology language.
\end{enumerate}
The OntoIOp activity builds on the results of a number of past standardisation  projects with best practices concerning structured content development and maintenance as well as with respect to content integration.  DOL, the Distributed Ontology Language, enables integration of a number of existing and future ontology languages (thanks to the registry mechanism), as well as ontology alignments and language/logic mappings, leading to sustainable interoperability among ontologies.  DOL's formal, machine-comprehensible semantics is provided by sophisticated state-of-the-art heterogeneous structuring mechanisms.  Grounding DOL's syntax on IRIs and giving it optional RDF and XML serializations affords reuse of existing annotation vocabularies to improve utility for services and comprehensibility for human users.

Future work will have a closer look at the interoperability of ontology-based services and devices.  Once there are standards, standards-based certification is possible. Especially with respect to eAccessibility \& eInclusion there is a definite need for certification, validation or verification of data, which possibly can largely be done through web services.  These standards and certification schemes would
\begin{enumerate}%[(1)]
\item first of all benefit end users (such as elderly people and persons with disabilities in the AAL use case introduced in section~\ref{sec:aal}),
\item benefit also small content and service providers,
\item be affordable, and
\item fit the kind of service, the technical state-of-the-art at the service providers' side and the expectations of the clients.
\end{enumerate}

% kwarc.bib contains all that I (Christoph) am interested in.  Let's add new references to paper.bib.  Additionally please copy any relevant *.bib files here.
\printbibliography
\end{document}